\documentclass[10pt,twocolumn,letterpaper]{article}

\usepackage{wacv}
\usepackage{times}
\usepackage{epsfig}
\usepackage{graphicx}
\usepackage{amsmath}
\usepackage{amssymb}
\usepackage{booktabs}
\usepackage{bm}
\usepackage{caption}

\def\wacvPaperID{1244} % Enter the WACV Paper ID here

\wacvalgorithmstrack   % Uncomment this line if you are submitting to the Algorithms Track.
\wacvfinalcopy % *** Uncomment this line for the final submission

\ifwacvfinal
\usepackage[breaklinks=true,bookmarks=false]{hyperref}
\else
\usepackage[pagebackref=true,breaklinks=true,colorlinks,bookmarks=false]{hyperref}
\fi

\begin{document}

%%%%%%%%% TITLE
\title{Placing Human Animations into 3D Scenes by Learning Interaction- and Geometry-Driven Keyframes}

\author{James F. Mullen Jr\\
University of Maryland\\
{\tt\small mullenj@umd.edu}
\and
Divya Kothandaraman\\
University of Maryland\\
{\tt\small dkr@umd.edu}
\and
Aniket Bera\\
Purdue University\\
{\tt\small ab@cs.purdue.edu}
\and
Dinesh Manocha\\
University of Maryland\\
{\tt\small dmanocha@umd.edu}
}

%\author {
%    % Authors
%    James F. Mullen Jr.\textsuperscript{\rm 1}, Divya Kothandaraman \textsuperscript{\rm 1}, Aniket Bera \textsuperscript{\rm 2}, Dinesh Manocha \textsuperscript{\rm 1} \\
%    \textsuperscript{\rm 1} University of Maryland, College Park, MD, USA\\
%    \textsuperscript{\rm 2} Purdue University, West Lafayette, IN, USA\\
%    %College Park, MD West Lafayette, IN\\
%    {\tt\small {mullenj,dkr,dmanocha}@umd.edu, ab@cs.purdue.edu}
%}

\twocolumn[{%
\renewcommand\twocolumn[1][]{#1}%
\maketitle

\vspace{-1.5em}
\begin{center}
    \centering
    \captionsetup{type=figure}
    \includegraphics[width=\textwidth]{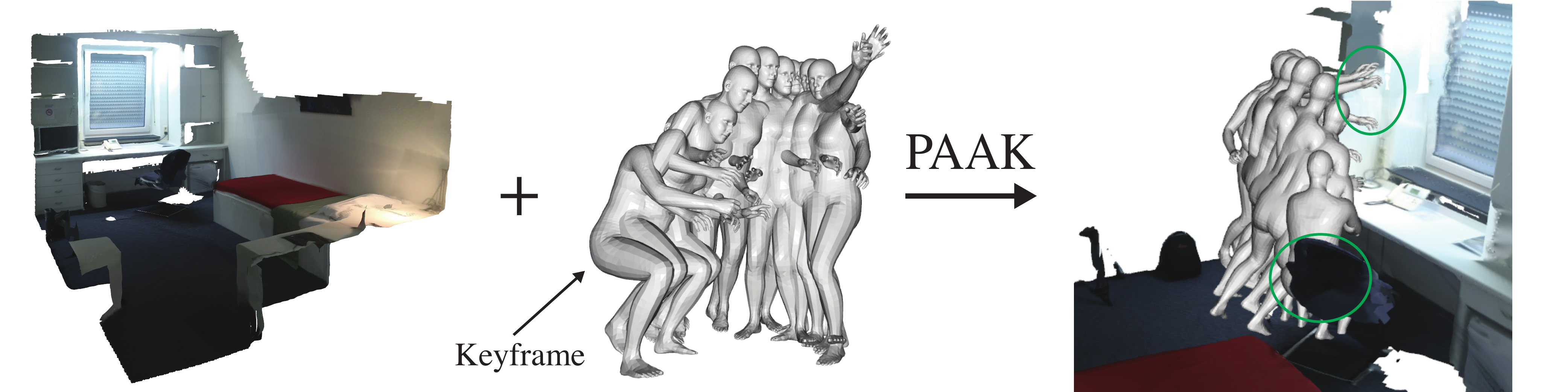}
    \captionof{figure}{Our goal is to place animations, a 3D sequence of human motion, into a 3D scene while maintaining any interactions with the scene the animation contains. First, we select ``keyframes," the most important meshes in the animation for modeling interactions with the scene. In the animation, the leftmost mesh where the human is sitting would be a keyframe. We then use the keyframes to find a placement in the scene that best matches the interactions in the animation (green circles, right).}
\end{center}%
}]
\thispagestyle{empty}

%%%%%%%%% ABSTRACT
\begin{abstract}
\vspace{-0.75em}
    We present a novel method for placing a 3D human animation into a 3D scene while maintaining any human-scene interactions in the animation. We use the notion of computing the most important meshes in the animation for the interaction with the scene, which we call ``keyframes." These keyframes allow us to better optimize the placement of the animation into the scene such that interactions in the animations (standing, laying, sitting, etc.) match the affordances of the scene (e.g., standing on the floor or laying in a bed). We compare our method, which we call PAAK, with prior approaches, including POSA, PROX ground truth, and a motion synthesis method, and highlight the benefits of our method with a perceptual study. Human raters preferred our PAAK method over the PROX ground truth data 64.6\% of the time. Additionally, in direct comparisons, the raters preferred PAAK over competing methods including 61.5\% compared to POSA. Our project website is available at \url{https://gamma.umd.edu/paak/}.
\end{abstract}

%%%%%%%%% BODY TEXT
\section{Introduction}

Throughout daily life, humans interact with their environment by making contact with objects and avoiding collisions with obstacles. Imagine you are sitting at your desk. Your arms may be resting on the desk. When you stand up to leave, you set your hands on the chair and desk as you walk around them. To leave the room you grab a doorknob to open the door. Interactions like these define how humans move throughout their environment. In this paper, our goal is to utilize these interactions to place animations into a scene in the most natural way.
%Consider a simple scenario where a human agent is sitting on a chair with their arms resting on a desk. When the human agent stands up to leave, they first set their hands on the chair and desk as they walk around the furniture. To leave the room, the human agent grabs a door knob to open the door. 

Applications in synthetic data generation, virtual reality (VR), augmented reality (AR)\cite{hassanPopulating3DScenes2021}, game design \cite{martinHowMakeImmersive} and human-robot interaction \cite{linVirtualRealityPlatform2016} need to consider interactions between humans and the environment while placing 3D human animations into 3D scenes. For example, an AR designer may want to populate a living room environment with people sitting on chairs or couches, navigating the space, and having a conversation in the corner. Current methods are based on animators using modeling or animation tools to generate such sequences, but this can be very time consuming and it requires animators with considerable experience using these tools.

%\textcolor{blue}Synthetic dataset creation can especially benefit from realistically placing human motion into a scene.  Since real world data contains humans interacting with their environments, synthetic data that does not model these interactions will miss key information that recognition models can exploit. Utilizing the affordances of a scene when integrating a human animation will allow the creation of more effective data for training action recognition systems.

%Prior methods for placing human actions into 3D scenes consider humans and scenes separately, where the human pose is typically estimated in isolation from the scene and 3D scenes are almost exclusively scanned and reconstructed without people. %For instance, parametric 3D human models such as SMPL \cite{loperSMPLSkinnedMultiperson2015} and SMPL-X \cite{pavlakosExpressiveBodyCapture2019} can represent human bodies accurately as 3D meshes but do not contain any information about the bodies interacting with the scene.

Prior work on human-scene interaction, or scene affordances, attempts to place 3D human models into 3D scene scans such that the placement matches real human behavior. Some methods focus on placing existing static human models into the scene \cite{kimShape2PoseHumancentricShape2014, kangEnvironmentAdaptiveContactPoses2014, hoSpatialRelationshipPreserving2010, hassanPopulating3DScenes2021} while others attempt to generate a suitable static human model\cite{zhangGenerating3DPeople2020, zhangPLACEProximityLearning2020, hassanResolving3DHuman2019}. However, many of these methods typically do not generalize well to new scenes and none work with 3D human animations. A recent method, POSA \cite{hassanPopulating3DScenes2021}, utilizes a cVAE \cite{sohnLearningStructuredOutput2015} to encode contact probabilities and semantic labels onto the vertices of a human model which can then be used to place it in the scene convincingly. This approach enables the generalization of the human models to any possible scene. POSA, and most other human-scene interaction methods, exclusively work with static, single-pose humans, not animations as we are interested in with our work.

%Recent work has made big strides in human-scene interaction \cite{gleicherRetargettingMotionNew1998, kimShape2PoseHumancentricShape2014, kangEnvironmentAdaptiveContactPoses2014, hoSpatialRelationshipPreserving2010, zhangGenerating3DPeople2020, hassanPopulating3DScenes2021, zhangPLACEProximityLearning2020, hassanResolving3DHuman2019}, placing 3D human models into the 3D scene scans. One such method, POSA \cite{hassanPopulating3DScenes2021}, learns a VAE \cite{sohnLearningStructuredOutput2015} to encode contact probabilities and semantic labels onto the vertices of a human model which can then be used to place it in the scene convincingly. This body-centric approach enables the generalization of the human models to any possible scene. POSA and other human-scene interaction methods almost exclusively work with static, single-pose humans. \textcolor{blue}{Divya: Issue with POSA that we tackle}

%\divya{Commenting this out, orthogonal to our work} Another line of active research focuses on human motion synthesis \cite{kovarFlexibleAutomaticMotion2003, harveyRobustMotionInbetweening2020, xiaLearningBasedSphereNonlinear2019, cleggLearningDressSynthesizing2018, wangSynthesizingLongTerm3D2021, wangDiverseNaturalSceneAware2022}. The animations these methods create are rapidly improving but continue to lack the realism of motion-captured human actions. Additionally, these methods rarely address the interactions between the humans and the scene. 

\subsection{Main Contributions}

We propose a novel method to \textit{place an existing 3D human animation into any arbitrary, static, 3D scene with natural-looking interactions}. We can use any human animations that are, or can be represented as, a time-series of SMPL-X \cite{pavlakosExpressiveBodyCapture2019} 3D human meshes and no assumptions are made about the scene itself. The scene can contain any arbitrary number of objects of varying shapes and at any location. Both the 3D animation and scene are inputs to our method which outputs the location and orientation of the most natural placement of the animation in the scene. The contributions of this paper are as follows:

\begin{enumerate}
    \item We present a deep learning-based selection method to find ``keyframes," frames in the animation most important for modeling a relevant scene interaction. Our method utilizes a deep model to determine potentially important frames from the animation's estimated interactions and geometry. Inspired by techniques from active learning literature we then calculate a diversity score for each mesh in the animation. Using the output of our deep model and the diversity score, we weight meshes in the animation such that the highest weight is attributed to meshes that maximize diversity and contain interactions with the scene important for a natural placement. We call the frames or meshes in the animation with the highest weight ``keyframes."
    
    \item We present an algorithm that utilizes the keyframes alongside a 3D scene's semantic information and affordances to optimize the placement of the animation into the scene. Our algorithm searches for target animation placements, optimizing the animations' position and orientation while maximizing the match between the estimated animation interactions and the scene geometry and semantics. We call our complete method PAAK, for ``Placement of Animations with Active Keyframes". 
    
    \item We qualitatively show natural and physically plausible human placement results. Through a perceptual study, we show that human raters prefer our method over PROX ground truth \cite{hassanResolving3DHuman2019} and rate it as more realistic than an extension of POSA to the time dimension (61.5\% v. 38.5\%), a generative method \cite{wangSynthesizingLongTerm3D2021} (76.9\% vs 23.1\%), and a purely geometric keyframe extraction method (52.3\% vs 47.7\%).
\end{enumerate}

%\divya{Implementation information shift to section 3} We first leverage POSA \cite{hassanPopulating3DScenes2021} for its body-centric contact and semantic labeling and annotate segments of the PROX dataset \cite{hassanResolving3DHuman2019}. We then utilize a highly customized version of BADGE, a prominent active learning method, alongside geometric and semantic equations, to find "keyframes," points in the animations most important for modelling the relevant scene interaction. 

%\divya{Shift to section 3} To visualize this process, imagine an animation where the human is sitting. It is important that at that frame they are sitting on a chair in the scene, however, if later in this animation the human stands up and walks forward, it is also important to select some frames during the motion such that the motion does not penetrate other objects in the scene.
\section{Related Work}
\textbf{Human Models.}
Most existing work utilizes body skeletons \cite{ionescuHuman36MLarge2014, sigalHumanEvaSynchronizedVideo2010} to model 3D humans. However, the surface of the body is important for rendering an actual human or modeling interactions with the environment or objects. Learned parametric 3D body models have addressed this need \cite{anguelovSCAPEShapeCompletion2005, jooTotalCapture3D2018, loperSMPLSkinnedMultiperson2015, pavlakosExpressiveBodyCapture2019, osmanSTARSparseTrained2020}. In this work, we utilize SMPL-X \cite{pavlakosExpressiveBodyCapture2019}, an extension of \cite{loperSMPLSkinnedMultiperson2015} that models face and hand articulation in addition to the rest of the body.

\textbf{Motion Synthesis.}
Motion synthesis is a longstanding problem in computer vision and computer graphics \cite{kovarFlexibleAutomaticMotion2003, pavlovicLearningSwitchingLinear2000, urtasunTopologicallyconstrainedLatentVariable2008, harveyRobustMotionInbetweening2020, starkeNeuralStateMachine2019, lingCharacterControllersUsing2020, xuHierarchicalStylebasedNetworks2020, holdenDeepLearningFramework2016, rempeHuMoR3DHuman2021, wangSynthesizingLongTerm3D2021, wangDiverseNaturalSceneAware2022}. Much of the early work on motion synthesis focused on synthesizing intermediate states between two given frames \cite{urtasunTopologicallyconstrainedLatentVariable2008, harveyRobustMotionInbetweening2020, xiaLearningBasedSphereNonlinear2019}. However, these methods are not able to handle large translational position changes effectively. Xu et al. \cite{xuHierarchicalStylebasedNetworks2020} and Holden et al. \cite{holdenDeepLearningFramework2016} utilized data-driven deep models for motion synthesis, showing better generalization than the geometric methods of the past. While many of these methods can create convincing 3D human motion, none of them address interactions between humans and scenes.

Some motion synthesis methods do consider both motion and the environment \cite{starkeNeuralStateMachine2019, lingCharacterControllersUsing2020, cleggLearningDressSynthesizing2018, wangCriticRegularizedRegression2020}. However, these methods use a greatly simplified scenario with predefined objects and primitive motion. Our work is closest to \cite{wangSynthesizingLongTerm3D2021} which accounts for an arbitrary 3D scene when synthesizing motion. The motions \cite{wangSynthesizingLongTerm3D2021} and other motion synthesis methods produce falls short of the realism of animations created through motion-capture. Our method does not synthesize its own motion and instead leverages motion capture data for the most realistic animation-scene pairing possible.

\begin{figure*}[ht]
	\begin{center}
		\includegraphics[width=.9\linewidth]{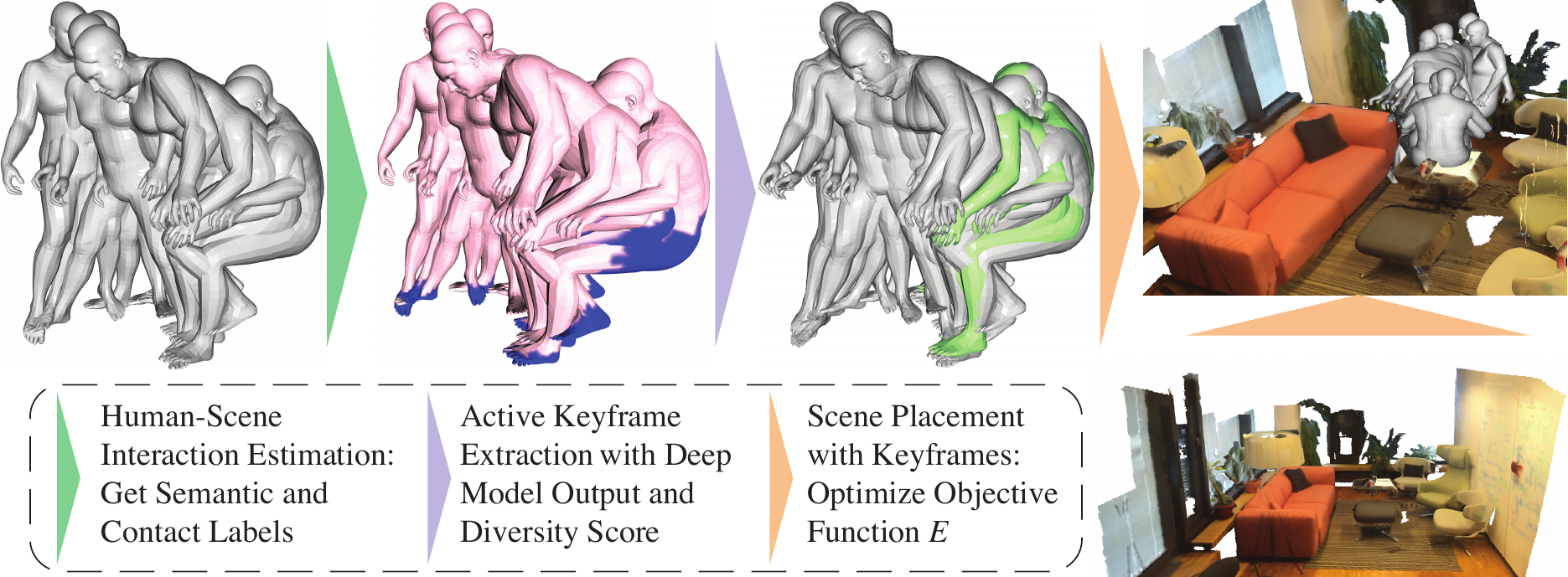}
		\caption{An overview of our PAAK method. We first estimate human-scene interactions and use those interactions to determine the keyframes in the animation. We can then utilize the keyframes alongside the 3D scene itself to place the animation convincingly into the scene.}
		\label{fig:method}
	\end{center}
\end{figure*}

\textbf{Video Synthesis.}
Our work is also related to the synthesis of full videos containing human actions. The advent of generative adversarial networks (GANS) \cite{goodfellowGenerativeAdversarialNets2014} and neural radiance fields (NeRF) \cite{mildenhallNeRFRepresentingScenes} have contributed to a growing body of work attempting to generate videos of humans completing actions in arbitrary scenes. Niemeyer et al. \cite{niemeyerGIRAFFERepresentingScenes2021} proposed a fully generative method which creates a full 3D scene with an arbitrary number of objects before rendering the scene into a 3D image. \cite{pengAnimatableNeuralRadiance2021, noguchiNeuralArticulatedRadiance2021, yangBANMoBuildingAnimatable2022, su2021anerf} worked towards extending NeRF to articulated objects like humans.

\cite{yangViSERVideoSpecificSurface2021, guSTYLENERFSTYLEBASED3DAWARE2022, zhaoHumanNeRFEfficientlyGenerated2022, kwonNeuralHumanPerformer2021, Chan2021, noguchiUnsupervisedLearningEfficient2022} extend these works towards full videos with human actions. Most closely related to our work is \cite{noguchiUnsupervisedLearningEfficient2022}, which similarly takes a human action and attempts to place it on a synthetic, usually sparse background. Our approach addresses the shortcomings of  \cite{noguchiUnsupervisedLearningEfficient2022}, especially the lack of natural-looking interactions between the animation and background and the lack of detail in the background. %Additional major differences between \cite{noguchiUnsupervisedLearningEfficient2022} and our method are our utilization of an existing 3D background instead of a generated one, our use of scene semantics and affordances when placing the animation into the scene, and our not utilizing GAN or NeRF techniques.

\textbf{Active Learning.}
%While a detailed literature review on active learning is not within the scope of this paper, we succinctly describe the idea and its applications. 
Active learning aims to annotate a small subset of a dataset to address a lack of sufficient labeled data. To do so, it computes the relevance of each sample based on a variety of parameters such as uncertainty/ entropy, diversity \cite{ashDEEPBATCHACTIVE2020}, and a careful trade-off of uncertainty and diversity \cite{prabhu2021active}. Its applications include object detection \cite{roy2018deep}, domain adaptation \cite{su2020active,kothandaraman2022distilladapt} and video tracking \cite{vondrick2011video}. In this paper, we aim to assign weights to frames in accordance with their relevance to potential human-scene interactions. For this, our algorithm computes the relevance of each frame by using a heuristic that is inspired by BADGE \cite{ashDEEPBATCHACTIVE2020}.

\textbf{Human-Scene Interaction.}
The main focus of our work is human-scene interaction (HSI), or scene affordance. Early work in HSI was purely geometric with Gleicher \cite{gleicherRetargettingMotionNew1998} using contact constraints for motion retargeting and Kim et al. \cite{kimShape2PoseHumancentricShape2014} automating the generation of 3D skeletons into a 3D environment. Subsequent geometric works continued to exploit the importance of contact and began to account for the forces present in the environment \cite{kangEnvironmentAdaptiveContactPoses2014, leimerPoseSeatAutomated2020, grabnerWhatMakesChair2011}. Gupta et al. \cite{gupta3DSceneGeometry2011} estimated the human poses "afforded" by the scene by predicting a 3D scene occupancy grid and computing the support and penetration of a 3D skeleton inside it. A subset of this research began to focus on dynamic interactions, or animations \cite{al-asqharRelationshipDescriptorsInteractive2013, hoSpatialRelationshipPreserving2010}.
 
More recently, data-driven approaches have begun to dominate \cite{tan2018, jiangHallucinatedHumansHidden2013, savvaSceneGrokInferringAction2014, zhangGenerating3DPeople2020, hassanPopulating3DScenes2021, liPuttingHumansScene2019}. Jiang et al. \cite{jiangHallucinatedHumansHidden2013} and Koppula et al. \cite{koppulaLearningHumanActivities2013} learn to estimate human poses and object affordances from an RGB-D 3D scene. Wang et al. \cite{wangGeometricPoseAffordance2021} learns to utilize the scene affordances to optimize pose estimation. Closest to our method, PSI \cite{zhangGenerating3DPeople2020}, PLACE \cite{zhangPLACEProximityLearning2020}, and POSA \cite{hassanPopulating3DScenes2021} populate scenes with SMPL-X \cite{pavlakosExpressiveBodyCapture2019} human meshes. POSA \cite{hassanPopulating3DScenes2021} is unique in its human-centric approach and utilization of dense body-to-scene contact. Specifically, in POSA, Hassan et al. uses a cVAE to learn contact probabilities and a corresponding semantic label for every vertex on a SMPL-X human mesh before using this information to find the best affordance for that mesh in a given 3D scene. In our method, we leverage POSA's model and utilize the contact and semantic information it provides. The key difference between our method and POSA is our use of an animation instead of a singular human mesh, creating additional challenges addressed by our keyframe methods.

\section{Placement of Human Animations into 3D Scenes}
In this work, we consider the following problem statement: \textit{Given a 3D human animation and a 3D scene, find the most natural-looking and physically plausible placement in the scene.} Specifically, our goal is to take a given animation and place it into a scene mesh such that any interactions in the animation (i.e. sitting in a chair, laying on a bed, or touching an object) match the affordances of the scene. The crux of our work is the idea that some frames in an animation are more important than others when optimizing the placement of an animation into a scene in the most natural way. Intuitively, you can imagine an animation with many static frames before a motion begins. When placing this animation into the scene treating all frames equally, the static frames will outweigh the moving ones in an optimizer as there are many more of them. In contrast, our method will find the moving frames and emphasize them to the optimizer, resulting in a more natural looking placement in the scene.

We present an overview of our method in \hyperref[fig:method]{Figure 2}. A list of symbols frequently used in this work is shown in \hyperref[table:symbols]{Table 1}. Rarely used symbols are defined where they are used. Note that notation relating to a time-series beginning with an uppercase letter denotes the full time-series while a lowercase letter denotes an individual frame in the time-series. We begin with a set of 3D human animations and a set of 3D scene meshes. Each 3D human animation, $V_b$ consists of a time series of human meshes, $v_b$, and skeletons represented by the SMPL-X\cite{pavlakosExpressiveBodyCapture2019} body model. We first estimate likely human-scene interactions given the poses of the human meshes in the animation. We then use a deep model to extract geometrically and semantically important keyframes before employing modified active learning techniques that leverage the deep model to extract a diversity score for each frame. The model output and diversity score are then combined to create ``Active Keyframes," $k$, frames in the animation that maximize diversity \textit{and} represent important interactions with the scene (shown as the green meshes in \hyperref[fig:method]{Figure 2}). Using our keyframe estimation we weigh the frames in the animation by their importance and use an optimizer to place the 3D human animation into the scene in the most natural way. This process results in a animation-scene pairing where interactions in the animation itself are matched with the geometry and semantics of the scene.
%by training a deep model to use these estimated interactions alongside the geometry of the meshes to emulate a geometric importance measure. This model enables us to calculate a diversity score for each mesh and combining this score with the geometric score results in our "active keyframes,"

\begin{table}[t] \label{table:symbols}
    \centering
    \begin{tabular}{| p{0.17\columnwidth} | p{0.73\columnwidth} | } 
        \hline
        \textbf{Symbols} & \textbf{Definitions}\\  
        \hline
        $V_b$ & A 3D human animation, consisting of a time series of 3D human meshes \\ 
        \hline
        $v_b$ & A single 3D human mesh, an individual frame from $V_b$ \\ 
        \hline
        $f_c$, $f_s$ & Contact labels and Semantic labels, attributed to each vertex in a mesh, $v_b$  \\ 
        \hline
        $K_g$, $K_a$ & Geometric and Active Keyframe weights respectively \\ 
        \hline
        $W_s$, $W_m$, $W_d$ & Semantic, motion, and diversity weights for an animation respectively\\
        \hline
        $E$ & The objective function utilized when optimizing the animations placement \\
        \hline
    \end{tabular}
    \caption{List of symbols used and their definitions.}
\end{table}

We chose an optimization method instead of end-to-end deep learning because it will \textit{always} find the ideal placement in the scene given the information we supply. In contrast, a deep model would learn to generalize over all animations and scenes, not necessarily finding the ideal placement for any of them. Our optimization method allows for the combination of our keyframe weights and individual meshes in the animation to prioritize interactions key to realism.

\subsection{Human-Scene Interaction Estimation}
To place an animation into the scene in a way that preserves any interactions present in the animation, it is essential to first determine what those interactions are. To encode the estimated relationship with a scene into the animation, we directly implement the POSA \cite{hassanPopulating3DScenes2021} model and feed in each frame of the animation individually to extract semantic and contact labels. POSA uses a conditional variational autoencoder (cVAE), $f$, to generate an egocentric feature map from each SMPL-X human mesh vertices in the animation. For example, when the mesh is of a person sitting in a chair, a vertex on a person's back should have a high contact probability and activate the chair semantic label, while a shin vertex would have a very low contact probability. POSA can be represented as the function $f$ in \hyperref[eq:POSA]{Equation 1}:
\begin{equation}\label{eq:POSA}
    f : (v_b) \rightarrow [f_c,f_s]
\end{equation}

%We chose to implement POSA because it provided all the semantic and contact information our keyframe extraction method required and has already been shown to work for placing static meshes of humans from the PROX \cite{hassanResolving3DHuman2019} dataset into scenes from the PROX dataset. An advantage of using POSA to extract semantic and contact information from an animation is the ability to find an arbitrary number of important interactions over an arbitrarily long animation. 
%Short-term temporal inconsistency with the POSA estimates was an issue we resolved by finding a dominating interaction for the animation. 

\subsection{Geometric Keyframes}
We present a novel geometric algorithm for placing a 3D animation into a scene such that interactions match the affordances of the scene given a 3D human animation, its likely interactions, and the scene. Optimizing the placement of the animation into the scene while weighting all the frames in the animation equally will miss important cues only present for a few of the frames. Weighting the frames higher if there is a likely interaction makes it much harder for the placement optimization process to miss key interactions. Semantic, contact, and geometric information of a frame relative to the animation itself are strong indicators of whether a given frame is essential for matching interactions in the animation with the affordances of the scene.

Our geometric keyframe weighting formula, $K_g$, \hyperref[eq:geom]{Equation 2}, consists of a weighted combination of semantic weights and motion weights. \hyperref[eq:ws]{Equation 3} and \hyperref[eq:wm]{Equation 4} show how the semantic term and motion term are calculated, respectively. Semantic weights are calculated for each mesh in the animation individually, where each individual weight sums the number of vertices in the mesh that are activated by the dominant semantic class in the animation, defined by the mode of the animation's semantic labels\footnote{excluding the floor class as it would almost always dominate}, $Mo(F_s)$. 

The motion weight uses the skeleton from the SMPL-X model and models lateral motion by taking the Euclidean distance from the pelvis at a frame, $p_i$, to the pelvis in the next frame, $p_{i+1}$. This provides a higher weight to frames with a quick motion than those where the human is static.

\begin{equation}\label{eq:geom}
    K_g = \lambda_s * \frac{W_s}{max(W_s)} + \lambda_m * \frac{W_m}{max(W_m)}
\end{equation}
\begin{equation}\label{eq:ws}
    w_s = \sum_{i=0}^{|v_b|}{[f_{s,i} = Mo(F_s)]}
\end{equation}
\begin{equation}\label{eq:wm}
    W_m = ||p_i - p_{i+1}||
\end{equation}

The $\lambda_s$ and $\lambda_m$ weighting factors are set through experimentation to achieve the best balance of semantic and motion weighting in the total keyframe weight.

Our Geometric Keyframes method can be thought of as prioritizing the meshes in the animation where an interaction with the scene is taking place or there is a large motion happening. For example, in an animation with a sitting action, the few frames where the human is sitting will be prioritized such that when placed into the scene, the human will be in a seat when sitting. Additionally, the motion weighting helps mitigate situations where large motions may cause the human to collide with obstacles in the environment when many static frames may otherwise dictate the placement. In practice, this leads to more natural-looking placements throughout the scene than with the semantic term alone.

\begin{figure}[t]
	\begin{center}
		\includegraphics[width=\columnwidth]{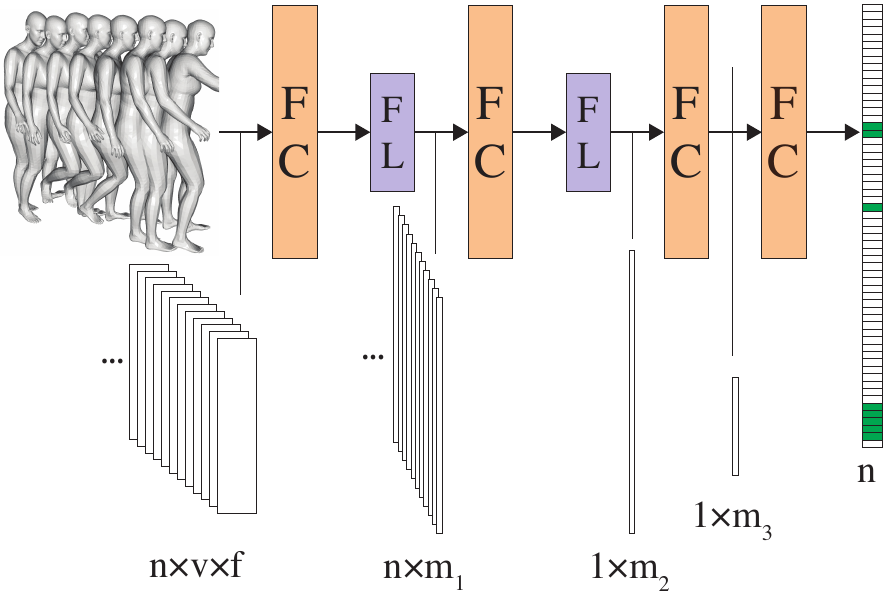}
		\caption{Network Architecture. The input animation is n meshes with v vertices and f features each. We utilize four fully connected (FC) layers with the first layer operating across each vertex while the second layer operates across all the vertices in the mesh. The last two layers operate across the entire animation. The model outputs an array of size n, with each index the weight of the corresponding mesh in the animation. The m values are intermediate representations and the FL layers correspond to a flattening of the input along the last two dimensions.}
		\label{fig:arch}
	\end{center}
\end{figure}

\subsection{Active Keyframes}
While our Geometric Keyframes method is effective at picking out important semantic cues and quick motions, we want to further improve it by increasing the diversity of the highly weighted frames and finding all important interactions. The increase in diversity will find frames ignored by our geometric formulation that may still be of relative importance when placing the animations into the scene. Our approach is motivated by active learning methods because while these try to find new pieces of an unlabeled dataset that would maximize diversity if annotated, we are looking for frames in an animation that would maximize the diversity of the high-weighted frames.

\begin{figure}[t]
	\begin{center}
		\includegraphics[width=0.9\columnwidth]{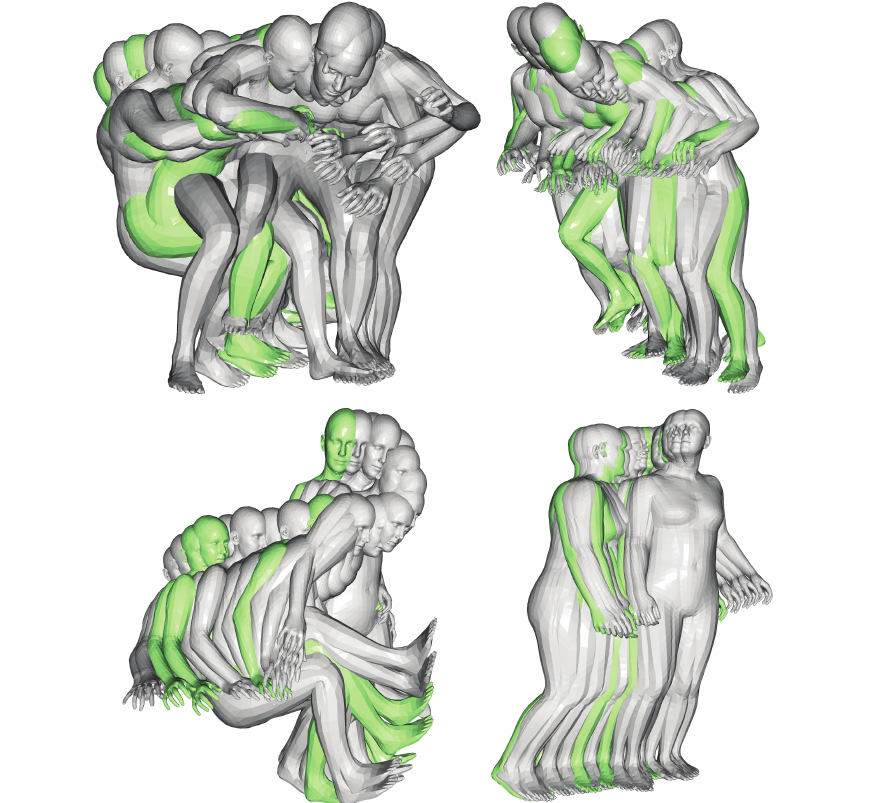}
		\caption{Random samples from our active keyframe framework. The green frames are those with the highest weight in $K_a$. Note that the frames where an important interaction occurs are preferred.}
		\label{fig:keyframes}
	\end{center}
\end{figure}

In our active keyframe method, $K_a$, \hyperref[eq:al]{Equation 5}, we begin with a deep fully-connected model. Our model can be thought of as a function, $g$, mapping from the animation vertices, contact labels, and semantic labels, to the geometric keyframe weights from \hyperref[eq:geom]{Equation 2}. The architecture of our model is shown in \hyperref[fig:arch]{Figure 3}. The primary benefit of this architecture is the ability to both connect through each mesh individually and across the animation. Another benefit of our model over our geometric equations is its generalization to multiple prominent interactions without forcing a set number of interactions to track. Utilizing a deep model also allows us to employ utilize techniques from active learning literature to compute a diversity score. Specifically, we use BADGE \cite{ashDEEPBATCHACTIVE2020} which uses the gradient of the model itself to determine a subset of the animation meshes that will maximize diversity. 

\begin{equation}
\label{eq:al}
    K_a = \lambda_g * \hat{K_g} + \lambda_b * W_d
\end{equation}
\begin{equation}\label{eq:model}
    g : (V_b, F_c, F_s) \rightarrow \hat{K_g}
\end{equation} % I used a hat here because its not actually k_g. Not sure how to reflect this in the text

To calculate the diversity score, BADGE divides samples into batches using kNN clustering on the gradient score. Samples are selected using both the magnitude of the gradient as well as the distance from previously selected samples. Since our goal is to obtain gradient-based diversity scores for each mesh rather than selecting samples for annotation, we run BADGE over all meshes/ frames of a video by setting the number of annotation frames to be equal to the number of frames. BADGE returns the gradient diversity score $w_d$ for each mesh in the animation.

\begin{figure*}[ht]
	\begin{center}
		\includegraphics[width=\linewidth]{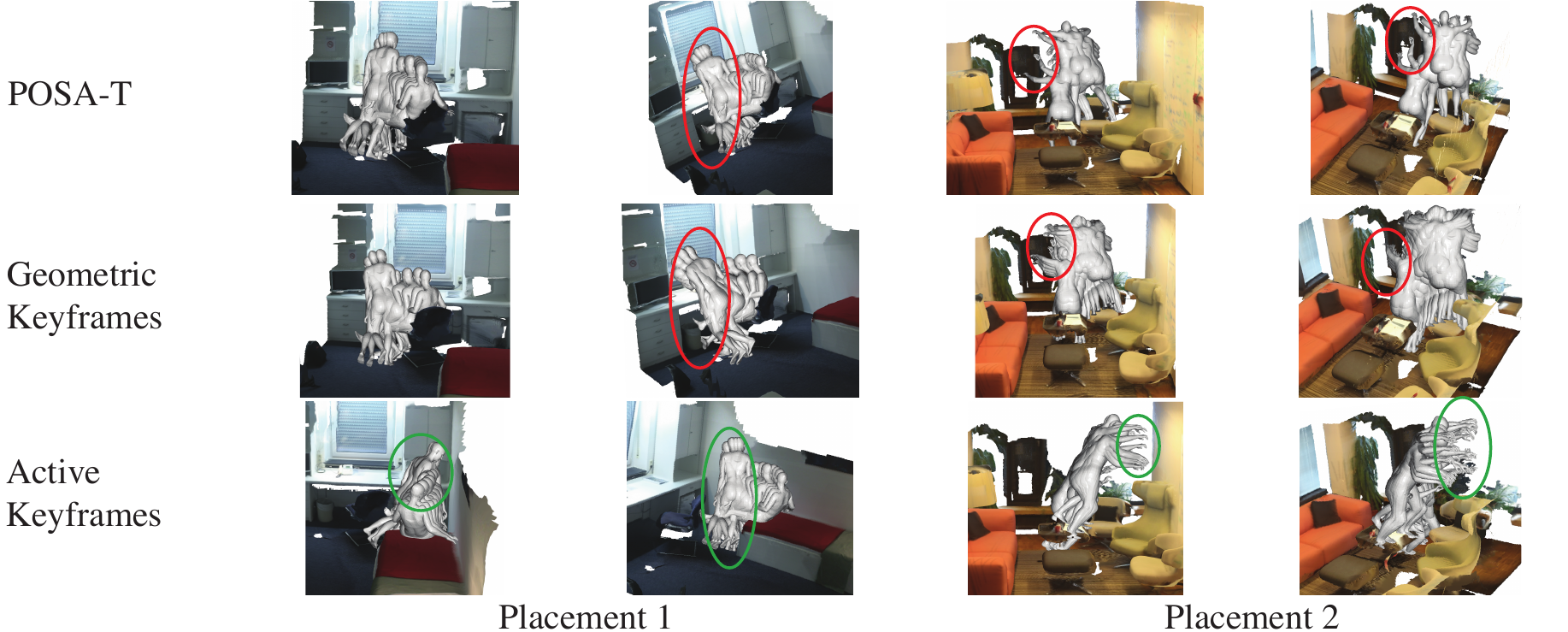}
		\caption{Comparisons on placing the same animation into the same scene across the POSA-T, Geometric Keyframes, and Active Keyframes Methods. Note that two angles of each placement are provided. For Placement 1, only the Active Keyframes method placed the animation on the bed (green circle), allowing for a more reclined seating position that results in standing more upright at the end of the animation. For Placement 2, a jumping action is taking place. Only the Active Keyframes method was able to position the animation such that the hands were not in collision with the back wall.}
		\label{fig:qual-comp}
	\end{center}
\end{figure*}

\subsection{Scene Placement with Keyframes}
Now that we have a keyframe weighting for the animation, we place the animation into the 3D scene such that it makes sense in the context of the scene, completing our PAAK method. Our approach takes the 3D scene mesh, the animation, and our keyframe weights and uses them to optimize an objective function, $E$. Optimizing $E$ finds the body translation in the scene, $\tau$, and the global body orientation, $\theta$, that minimize the sum of an affordance loss, $\mathcal{L}_{afford}$, and a penetration loss, $\mathcal{L}_{pen}$, calculated for each frame weighted by our keyframe weights, $k_g$ or $k_a$.
\begin{equation}
    E(\tau, \theta) = \sum^{|k|}_{i=0}{k_{i}*[\mathcal{L}_{afford, i}} + \mathcal{L}_{pen, i}]
\end{equation}

Both $\mathcal{L}_{afford}$ and $\mathcal{L}_{pen}$ are adapted from \cite{hassanPopulating3DScenes2021}.
$\mathcal{L}_{afford}$ is minimized when the distance to the scene is small for vertices with a high probability of contact, $f_c$, and when the semantic label, $f_s$, matches the semantics of the object it is touching in the scene. $\mathcal{L}_{pen}$ heavily punishes a placement that results in a mesh penetrating the scene. By weighting these values with our keyframe weights, we make sure that the minimum value of $E$ is found by maximizing correct affordances and minimizing penetration at the Keyframes. Without our keyframe weighting, the objective function becomes saturated across the meshes in the animation, not placing the animation such that it matches the scene context adequately. Optimizing $E$ without our keyframe weighting is a baseline we use in our experiments.

\section{Experiments}

% To supplementary \subsection{Implementation Details}
%We chose to use 2-second animations, which gives us 60 frames per animation at 30 fps. To train, we used the Adam optimizer, a learning rate of 0.00001, and a batch size of 1. We trained for 5 epochs across the PROX dataset. Our $\lambda$ weight values are set as follows: $\lambda_s = \lambda_g = \lambda_b = 1$, $\lambda_m = 0.2$. 

\subsection{Dataset and Baselines}
For all of our experiments, we utilized the PROX dataset \cite{hassanResolving3DHuman2019}, which contains animations of humans moving through 12 different scenes. The PROX ground-truth (GT) SMPL-X \cite{pavlakosExpressiveBodyCapture2019} parameters are generated by a fitting algorithm, introducing some noise into the animations. From PROX, we sample 10k animations from 8 of the 12 available scenes. For evaluation, we randomly sample animations from our PROX subset and place them into one of the four remaining scenes. These results are then shown to human raters in comparison with PROX GT or another method, and these raters then pick the more realistic of two videos. Both videos are from the same scene.

We compare our method with POSA and a motion synthesis method as baselines. Additionally, we ablate our Active Keyframes method with our Geometric Keyframes method. The baselines are outlined in further detail below.%The motion synthesis method \cite{wangSynthesizingLongTerm3D2021} is a baseline to help further frame the problem of naturally placing animations in 3D scenes. %This also follows the precedent in \cite{hassanPopulating3DScenes2021} of comparing a placement method with a fully generative one.

\textbf{POSA-T.} Hassan et al. \cite{hassanPopulating3DScenes2021} propose a method that places a single human body into a scene given its affordances. As this approach does not consider a full animation we sum the loss over all the meshes in the animation and provide that to the optimizer. We call this baseline POSA-T for our addition of the time dimension.

\textbf{Motion Synthesis.} Wang et al. \cite{wangSynthesizingLongTerm3D2021} propose a motion synthesis method that accounts for the affordances of the scene in its motion creation. %Their two-stage model uses a cVAE to generate static poses at the beginning and end of the animation before connecting them with a route and then poses along that route. 
We did not alter their approach and used it as designed with the same four scenes it set aside for testing. We call this baseline Motion Synthesis.

\textbf{Geometric Keyframes.} Our Geometric Keyframes method is an ablative baseline that does not include our model nor our active keyframe implementation. It extracts the keyframe weights purely from semantic and geometric information, as described in \hyperref[eq:geom]{Equations 2-4}. We call this baseline Geometric Keyframes.

\subsection{Evaluation}
A qualitative comparison of PAAK alongside the POSA-T and Geometric Keyframes baselines can be found in \hyperref[fig:qual-comp]{Figure 5}. Qualitatively, we found that PAAK calculated improved placements over the baselines, with POSA-T especially prone to producing a placement that was not valid in the scene context, like having the human sit in mid-air.

\textbf{Comparison to PROX ground truth.} Following the protocols of Hassan et al. \cite{hassanPopulating3DScenes2021} and Zhang et al. \cite{zhangPLACEProximityLearning2020}, we compare our results to randomly selected examples from PROX ground truth. We utilize 4 real 3D scenes from the PROX test dataset, namely MPH16, MPH1Library, N0SittingBooth, and N3OpenArea. We then take 80 2-second animations from the PROX training dataset (not from the 4 scenes listed) for placement into the test scenes with each of our methods. The placement process for every method begins with a grid of potential placement locations and orientations on the scene, with each placement testing rotations every 30 degrees. We then filter these initial placements to 10 promising prospects based on $E$ and continue to optimize them until each reaches its optimum translation and rotation. For each of these 10 prospects, the location with the lowest total loss is selected as the final placement location. We render each animation scene pair into videos from four different angles so subjects can get a sense of the relationships between the animation and the scene. Using a web-based user study with 18 subjects, each being shown a subset of the image scene pairs we produced, we collect 540 unique ratings. The results are shown in \hyperref[table:gt]{Table 2}. The Geometric Keyframes ablation is almost indistinguishable from the PROX GT, while POSA-T falls short. However, the human raters preferred PAAK over PROX GT. We believe this is due to scene penetrations in the PROX dataset when humans are sitting. This is caused by deformations in real life not captured in the scene.

\begin{table}[t] \label{table:gt}
    \centering
    \begin{tabular}{l c c } 
        \hline\hline
        & Placement $\uparrow$ & PROX GT $\downarrow$ \\ [0.5ex] 
        \hline\hline
        POSA-T & 44.6\% & 55.4\% \\ 
        \hline
        Geom. Keyframes & 52.3\% & 47.7\% \\
        \hline
        PAAK & \textbf{64.6\%} & 35.4\% \\
        \hline
    \end{tabular}
    \caption{Comparison to PROX \cite{hassanResolving3DHuman2019} ground truth. Subjects are shown pairs of an action placed into a 3D scene and PROX ground truth (GT) and must choose the most realistic one. A higher percentage indicates the scene subjects deemed more realistic.}
\end{table}

\begin{table}[t] \label{table:direct}
    \centering
    \begin{tabular}{l c c } 
        \hline\hline
        & Baseline $\downarrow$ & PAAK $\uparrow$ \\ [0.5ex] 
        \hline\hline
        POSA-T & 38.5\% & \textbf{61.5\%} \\ 
        \hline
        Motion Synthesis & 23.1\% & \textbf{76.9\%} \\
        \hline
        Geom. Keyframes & 47.7\% & \textbf{52.3\%} \\
        \hline
    \end{tabular}
    \caption{PAAK compared to POSA-T, Motion Synthesis \cite{wangSynthesizingLongTerm3D2021}, and our Geometric Keyframes ablation. The comparison procedure is the same as for \hyperref[table:gt]{Table 2}.}
\end{table}

\textbf{Comparison to baselines.} To compare the baselines directly, we follow the same protocol as above, but replace the PROX ground truth with a competing method. In addition to comparing against POSA-T and the Geometric Keyframes ablative baseline, we also directly compare against Motion Synthesis. The results are shown in \hyperref[table:direct]{Table 3}. Again, we find that the human raters preferred PAAK, finding it more realistic than the other methods. PAAK significantly outperforms the Motion Synthesis baseline. This makes sense as the Motion Synthesis method is solving a slightly different task, generating natural-looking motion, while our method utilizes real-world data. This shows the need for a method like ours that uses motion captured animations. %The Motion Synthesis method does have the advantage of being able to produce an animation that specifically fits the given scene instead of trying to shoehorn one that might not truly fit. 
%The underperformance of the Motion Synthesis method, the current state of the art for motion synthesis accounting for the affordances in 3D scenes, proves the need for a method like ours that utilizes animations captured from the real-world.

\begin{figure}[t]
	\begin{center}
		\includegraphics[width=.9\columnwidth]{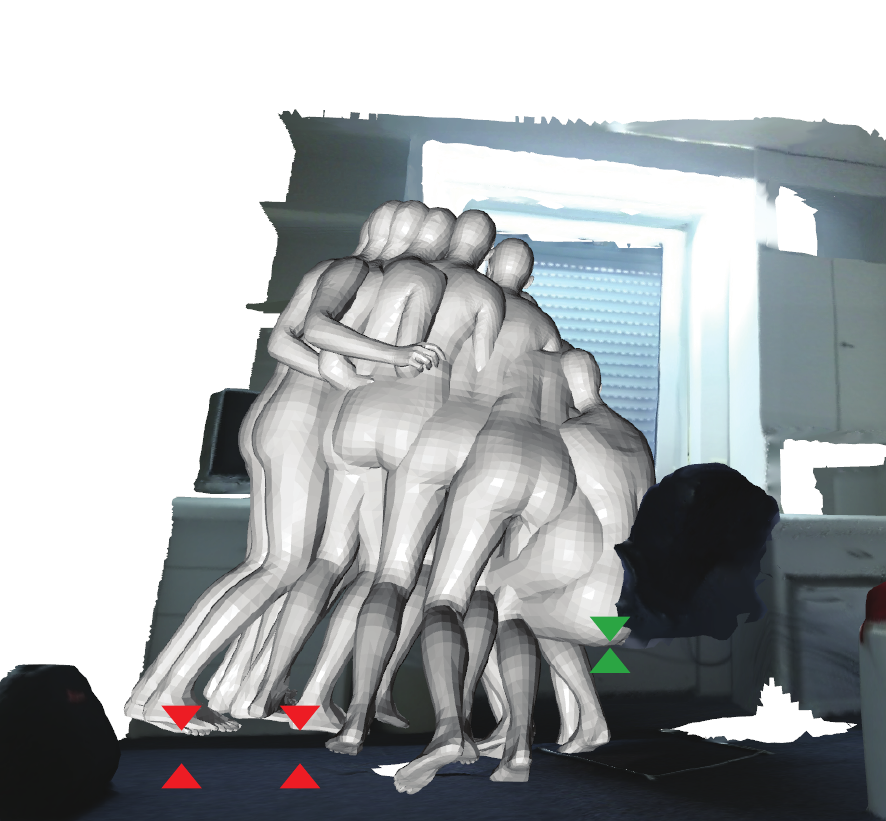}
		\caption{An example of a Keyframe-based placement leaving feet off the floor to place the buttocks correctly on the seat.}
		\label{fig:qual}
	\end{center}
\end{figure}

PAAK placements are also perceived by the human raters as more natural-looking than the POSA-T method. This makes sense as the POSA-T method will weigh all the individual meshes equally when placing the animation into the scene, making it more difficult for the optimizer to find a placement that maximizes the realism of the motion, like landing on a surface when sitting or touching an object when reaching. PAAK also outperforms our Geometric Keyframes ablative baseline. We believe this is due to the intelligent frame selection present in the method, which selects a more diverse set of keyframes to supplement the frames selected by the model for their semantic and geometric information. This added diversity can help to pick up on additional semantic cues not utilized in the Geometric Keyframes method.

\textbf{Physical plausibility.} Following the procedures utilized by \cite{hassanPopulating3DScenes2021, zhangPLACEProximityLearning2020, zhangGenerating3DPeople2020}, we take 100 animations of 60 frames each and place them in the 4 test scenes of PROX. Given the body meshes, the scene mesh, and a scene signed distance field (SDF), we compute a non-collision score and contact score as defined by \cite{zhangGenerating3DPeople2020} with the results in \hyperref[table:metric]{Table 4}. The non-collision score is calculated for each mesh in an animation as the ratio of the body vertices with positive SDF values divided by the total number of SMPL-X vertices. A high non-collision score denotes that the meshes in each animation do not penetrate the scene. PAAK is comparable to the POSA-T and Geometric Keyframes baselines in the non-collision score.

The contact score is calculated for each mesh individually and is 1 if at least one vertex of the mesh is in direct contact with the scene. PAAK is comparable to the POSA-T and geometric keyframes baselines in the contact score. The small difference in performance is likely due to slight mismatches between the animation and the scene. For example, in \hyperref[fig:qual]{Figure 6} PAAK ensures contact with a seat when sitting, however, an imperfect match in seat size between the animation and the scene results in feet close to but not completely in contact with the ground when the person stands. Comparatively, a placement that makes sure the feet are in contact with the ground for every mesh in the animation would get a perfect score. 

\begin{table}[t] \label{table:metric}
    \centering
    \begin{tabular}{l c c } 
        \hline\hline
        & Non-Collision $\uparrow$ & Contact $\uparrow$ \\ [0.5ex] 
        \hline\hline
        POSA-T & 0.98 & \textbf{0.83} \\ 
        \hline
        Geometric Keyframes & 0.98 & 0.81 \\
        \hline
        PAAK & \textbf{0.99} & 0.81 \\
        \hline
    \end{tabular}
    \caption{Evaluation of the physical plausibility metrics. A higher score is better for both.}
\end{table}
\section{Conclusions, Limitations, and Future Work}
In this paper, we propose PAAK, a novel method for placing a 3D human animation into a 3D scene with accurately modeled human-scene interactions. We introduce ``keyframes," the frames in an animation most important for the interactions with the scene, and use these keyframes to place the animation into a scene. Human raters preferred PAAK animation placements over real-world PROX \cite{hassanResolving3DHuman2019} ground truth data, and over existing methods.

\textbf{Limitations and Future Work.} Note that PAAK does not \textit{always} create natural placements. There are still cases where people sit in strange places or walk where they typically would not. Our optimization method can miss good placements as it relies on a grid of initial placements before optimizing the best ones. For example, initial placements with heavily penalized penetrations could become the best available with further optimization. We limited this for time but more compute could enable improved placements. The ability to rate the quality of a placement would be valuable for end users. For example, if an animator has a bank of 100 animations, how can they pick the top 5 to populate the scene?
PAAK can be extended to model human-human interactions when placing multiple animations into a scene by adding a term to the semantic keyframe extraction. Finally, altering the animation itself could be a valuable extension that allows for more natural-looking interactions with the scene and better performance in the physical plausibility metrics.
\newline\newline
\textbf{Acknowledgements.} This material is based upon work supported by the National Science Foundation Graduate Research Fellowship Program under Grant No. DGE 1840340. Any opinions, findings, and conclusions or recommendations expressed in this material are those of the author(s) and do not necessarily reflect the views of the National Science Foundation. This research was supported by Army Cooperative Agreement W911NF2120076.

{\small
\bibliographystyle{ieee_fullname}
\bibliography{egbib}

\begin{thebibliography}{10}\itemsep=-1pt

\bibitem{al-asqharRelationshipDescriptorsInteractive2013}
Rami~Ali {Al-Asqhar}, Taku Komura, and Myung~Geol Choi.
\newblock Relationship descriptors for interactive motion adaptation.
\newblock In {\em Proceedings of the 12th {{ACM SIGGRAPH}}/{{Eurographics
  Symposium}} on {{Computer Animation}}}, {{SCA}} '13, pages 45--53, {New York,
  NY, USA}, July 2013. {Association for Computing Machinery}.

\bibitem{anguelovSCAPEShapeCompletion2005}
Dragomir Anguelov, Praveen Srinivasan, Daphne Koller, Sebastian Thrun, Jim
  Rodgers, and James Davis.
\newblock {{SCAPE}}: Shape completion and animation of people.
\newblock {\em ACM Transactions on Graphics}, 24(3):408--416, July 2005.

\bibitem{ashDEEPBATCHACTIVE2020}
Jordan~T Ash, Chicheng Zhang, and Akshay Krishnamurthy.
\newblock {{DEEP BATCH ACTIVE LEARNING BY DIVERSE}}, {{UNCERTAIN GRADIENT LOWER
  BOUNDS}}.
\newblock {\em ICLR}, page~26, 2020.

\bibitem{Chan2021}
Eric~R. Chan, Connor~Z. Lin, Matthew~A. Chan, Koki Nagano, Boxiao Pan,
  Shalini~De Mello, Orazio Gallo, Leonidas Guibas, Jonathan Tremblay, Sameh
  Khamis, Tero Karras, and Gordon Wetzstein.
\newblock Efficient geometry-aware {{3D}} generative adversarial networks.
\newblock In {\em {{arXiv}}}, 2021.

\bibitem{cleggLearningDressSynthesizing2018}
Alexander Clegg, Wenhao Yu, Jie Tan, C.~Karen Liu, and Greg Turk.
\newblock Learning to dress: Synthesizing human dressing motion via deep
  reinforcement learning.
\newblock {\em ACM Transactions on Graphics}, 37(6):179:1--179:10, Dec. 2018.

\bibitem{gleicherRetargettingMotionNew1998}
Michael Gleicher.
\newblock Retargetting motion to new characters.
\newblock In {\em Proceedings of the 25th Annual Conference on {{Computer}}
  Graphics and Interactive Techniques - {{SIGGRAPH}} '98}, pages 33--42, {Not
  Known}, 1998. {ACM Press}.

\bibitem{goodfellowGenerativeAdversarialNets2014}
Ian Goodfellow, Jean {Pouget-Abadie}, Mehdi Mirza, Bing Xu, David
  {Warde-Farley}, Sherjil Ozair, Aaron Courville, and Yoshua Bengio.
\newblock Generative {{Adversarial Nets}}.
\newblock In {\em Advances in {{Neural Information Processing Systems}}},
  volume~27. {Curran Associates, Inc.}, 2014.

\bibitem{grabnerWhatMakesChair2011}
Helmut Grabner, Juergen Gall, and Luc Van~Gool.
\newblock What makes a chair a chair?
\newblock In {\em {{CVPR}} 2011}, pages 1529--1536, June 2011.

\bibitem{guSTYLENERFSTYLEBASED3DAWARE2022}
Jiatao Gu, Lingjie Liu, Peng Wang, and Christian Theobalt.
\newblock {{STYLENERF}}: {{A STYLE-BASED 3D-AWARE GENERA- TOR FOR
  HIGH-RESOLUTION IMAGE SYNTHESIS}}.
\newblock {\em ICLR}, page~25, 2022.

\bibitem{gupta3DSceneGeometry2011}
Abhinav Gupta, Scott Satkin, Alexei~A. Efros, and Martial Hebert.
\newblock From {{3D}} scene geometry to human workspace.
\newblock In {\em {{CVPR}} 2011}, pages 1961--1968, {Colorado Springs, CO,
  USA}, June 2011. {IEEE}.

\bibitem{harveyRobustMotionInbetweening2020}
F{\'e}lix~G. Harvey, Mike Yurick, Derek Nowrouzezahrai, and Christopher Pal.
\newblock Robust motion in-betweening.
\newblock {\em ACM Transactions on Graphics}, 39(4):60:60:1--60:60:12, July
  2020.

\bibitem{hassanResolving3DHuman2019}
Mohamed Hassan, Vasileios Choutas, Dimitrios Tzionas, and Michael Black.
\newblock Resolving {{3D Human Pose Ambiguities With 3D Scene Constraints}}.
\newblock In {\em 2019 {{IEEE}}/{{CVF International Conference}} on {{Computer
  Vision}} ({{ICCV}})}, pages 2282--2292, {Seoul, Korea (South)}, Oct. 2019.
  {IEEE}.

\bibitem{hassanPopulating3DScenes2021}
Mohamed Hassan, Partha Ghosh, Joachim Tesch, Dimitrios Tzionas, and Michael~J.
  Black.
\newblock Populating {{3D Scenes}} by {{Learning Human-Scene Interaction}}.
\newblock In {\em 2021 {{IEEE}}/{{CVF Conference}} on {{Computer Vision}} and
  {{Pattern Recognition}} ({{CVPR}})}, pages 14703--14713, {Nashville, TN,
  USA}, June 2021. {IEEE}.

\bibitem{hoSpatialRelationshipPreserving2010}
Edmond S.~L. Ho, Taku Komura, and Chiew-Lan Tai.
\newblock Spatial relationship preserving character motion adaptation.
\newblock {\em ACM Transactions on Graphics}, 29(4):33:1--33:8, July 2010.

\bibitem{holdenDeepLearningFramework2016}
Daniel Holden, Jun Saito, and Taku Komura.
\newblock A deep learning framework for character motion synthesis and editing.
\newblock {\em ACM Transactions on Graphics}, 35(4):138:1--138:11, July 2016.

\bibitem{ionescuHuman36MLarge2014}
Catalin Ionescu, Dragos Papava, Vlad Olaru, and Cristian Sminchisescu.
\newblock Human3.{{6M}}: {{Large Scale Datasets}} and {{Predictive Methods}}
  for {{3D Human Sensing}} in {{Natural Environments}}.
\newblock {\em IEEE Transactions on Pattern Analysis and Machine Intelligence},
  36(7):1325--1339, July 2014.

\bibitem{jiangHallucinatedHumansHidden2013}
Yun Jiang, Hema Koppula, and Ashutosh Saxena.
\newblock Hallucinated {{Humans}} as the {{Hidden Context}} for {{Labeling 3D
  Scenes}}.
\newblock In {\em Proceedings of the {{IEEE Conference}} on {{Computer Vision}}
  and {{Pattern Recognition}}}, pages 2993--3000, 2013.

\bibitem{jooTotalCapture3D2018}
Hanbyul Joo, Tomas Simon, and Yaser Sheikh.
\newblock Total {{Capture}}: {{A 3D Deformation Model}} for {{Tracking Faces}},
  {{Hands}}, and {{Bodies}}.
\newblock In {\em 2018 {{IEEE}}/{{CVF Conference}} on {{Computer Vision}} and
  {{Pattern Recognition}}}, pages 8320--8329, {Salt Lake City, UT, USA}, June
  2018. {IEEE}.

\bibitem{kangEnvironmentAdaptiveContactPoses2014}
Changgu Kang and Sung-Hee Lee.
\newblock Environment-{{Adaptive Contact Poses}} for {{Virtual Characters}}.
\newblock {\em Computer Graphics Forum}, 33(7):1--10, 2014.

\bibitem{kimShape2PoseHumancentricShape2014}
Vladimir~G. Kim, Siddhartha Chaudhuri, Leonidas Guibas, and Thomas Funkhouser.
\newblock {{Shape2Pose}}: Human-centric shape analysis.
\newblock {\em ACM Transactions on Graphics}, 33(4):120:1--120:12, July 2014.

\bibitem{koppulaLearningHumanActivities2013}
Hema~Swetha Koppula, Rudhir Gupta, and Ashutosh Saxena.
\newblock Learning human activities and object affordances from {{RGB-D}}
  videos.
\newblock {\em International Journal of Robotics Research}, 32(8):951--970,
  July 2013.

\bibitem{kothandaraman2022distilladapt}
Divya Kothandaraman, Sumit Shekhar, Abhilasha Sancheti, Manoj Ghuhan, Tripti
  Shukla, and Dinesh Manocha.
\newblock Distilladapt: Source-free active visual domain adaptation.
\newblock {\em arXiv preprint arXiv:2205.12840}, 2022.

\bibitem{kovarFlexibleAutomaticMotion2003}
Lucas Kovar and Michael Gleicher.
\newblock Flexible automatic motion blending with registration curves.
\newblock In {\em Proceedings of the 2003 {{ACM SIGGRAPH}}/{{Eurographics}}
  Symposium on {{Computer}} Animation}, {{SCA}} '03, pages 214--224, {Goslar,
  DEU}, July 2003. {Eurographics Association}.

\bibitem{kwonNeuralHumanPerformer2021}
Youngjoong Kwon, Dahun Kim, Duygu Ceylan, and Henry Fuchs.
\newblock Neural {{Human Performer}}: {{Learning Generalizable Radiance
  Fields}} for {{Human Performance Rendering}}.
\newblock In {\em Advances in {{Neural Information Processing Systems}}},
  volume~34, pages 24741--24752. {Curran Associates, Inc.}, 2021.

\bibitem{leimerPoseSeatAutomated2020}
Kurt Leimer, Andreas Winkler, Stefan Ohrhallinger, and Przemyslaw Musialski.
\newblock Pose to {{Seat}}: {{Automated}} design of body-supporting surfaces.
\newblock {\em Computer Aided Geometric Design}, 79:101855, May 2020.

\bibitem{liPuttingHumansScene2019}
Xueting Li, Sifei Liu, Kihwan Kim, Xiaolong Wang, Ming-Hsuan Yang, and Jan
  Kautz.
\newblock Putting {{Humans}} in a {{Scene}}: {{Learning Affordance}} in {{3D
  Indoor Environments}}.
\newblock In {\em 2019 {{IEEE}}/{{CVF Conference}} on {{Computer Vision}} and
  {{Pattern Recognition}} ({{CVPR}})}, pages 12360--12368, {Long Beach, CA,
  USA}, June 2019. {IEEE}.

\bibitem{linVirtualRealityPlatform2016}
Jenny Lin, Xingwen Guo, Jingyu Shao, Chenfanfu Jiang, Yixin Zhu, and Song-Chun
  Zhu.
\newblock A virtual reality platform for dynamic human-scene interaction.
\newblock In {\em {{SIGGRAPH ASIA}} 2016 {{Virtual Reality}} Meets {{Physical
  Reality}}: {{Modelling}} and {{Simulating Virtual Humans}} and
  {{Environments}}}, {{SA}} '16, pages 1--4, {New York, NY, USA}, Nov. 2016.
  {Association for Computing Machinery}.

\bibitem{lingCharacterControllersUsing2020}
Hung~Yu Ling, Fabio Zinno, George Cheng, and Michiel Van De~Panne.
\newblock Character controllers using motion {{VAEs}}.
\newblock {\em ACM Transactions on Graphics}, 39(4):40:40:1--40:40:12, July
  2020.

\bibitem{loperSMPLSkinnedMultiperson2015}
Matthew Loper, Naureen Mahmood, Javier Romero, Gerard {Pons-Moll}, and
  Michael~J. Black.
\newblock {{SMPL}}: A skinned multi-person linear model.
\newblock {\em ACM Transactions on Graphics}, 34(6):1--16, Nov. 2015.

\bibitem{martinHowMakeImmersive}
Jennifer Martin.
\newblock How to {{Make Immersive Game Design}} | {{University}} of {{Silicon
  Valley}}.
\newblock https://usv.edu/blog/how-to-make-immersive-game-design/, 2020.

\bibitem{mildenhallNeRFRepresentingScenes}
Ben Mildenhall, Pratul~P Srinivasan, Matthew Tancik, Jonathan~T Barron, Ravi
  Ramamoorthi, and Ren Ng.
\newblock {{NeRF}}: {{Representing Scenes}} as {{Neural Radiance Fields}} for
  {{View Synthesis}}.
\newblock {\em ECCV}, page~17, 2020.

\bibitem{niemeyerGIRAFFERepresentingScenes2021}
Michael Niemeyer and Andreas Geiger.
\newblock {{GIRAFFE}}: {{Representing Scenes}} as {{Compositional Generative
  Neural Feature Fields}}.
\newblock In {\em 2021 {{IEEE}}/{{CVF Conference}} on {{Computer Vision}} and
  {{Pattern Recognition}} ({{CVPR}})}, pages 11448--11459, {Nashville, TN,
  USA}, June 2021. {IEEE}.

\bibitem{noguchiUnsupervisedLearningEfficient2022}
Atsuhiro Noguchi, Xiao Sun, Stephen Lin, and Tatsuya Harada.
\newblock Unsupervised {{Learning}} of {{Efficient Geometry-Aware Neural
  Articulated Representations}}, Apr. 2022.

\bibitem{noguchiNeuralArticulatedRadiance2021}
Atsuhiro Noguchi, Sun Xiao, Stephen Lin, and Tatsuya Harada.
\newblock Neural {{Articulated Radiance Field}}.
\newblock In {\em Proceedings of the {{IEEE}}/{{CVF International Conference}}
  on {{Computer Vision}} ({{ICCV}})}, pages 5762--5772, Oct. 2021.

\bibitem{osmanSTARSparseTrained2020}
Ahmed A.~A. Osman, Timo Bolkart, and Michael~J. Black.
\newblock {{STAR}}: {{Sparse Trained Articulated Human Body Regressor}}.
\newblock In Andrea Vedaldi, Horst Bischof, Thomas Brox, and Jan-Michael Frahm,
  editors, {\em Computer {{Vision}} \textendash{} {{ECCV}} 2020}, volume 12351,
  pages 598--613. {Springer International Publishing}, {Cham}, 2020.

\bibitem{pavlakosExpressiveBodyCapture2019}
Georgios Pavlakos, Vasileios Choutas, Nima Ghorbani, Timo Bolkart, Ahmed~A.
  Osman, Dimitrios Tzionas, and Michael~J. Black.
\newblock Expressive {{Body Capture}}: {{3D Hands}}, {{Face}}, and {{Body
  From}} a {{Single Image}}.
\newblock In {\em 2019 {{IEEE}}/{{CVF Conference}} on {{Computer Vision}} and
  {{Pattern Recognition}} ({{CVPR}})}, pages 10967--10977, {Long Beach, CA,
  USA}, June 2019. {IEEE}.

\bibitem{pavlovicLearningSwitchingLinear2000}
Vladimir Pavlovic, James~M Rehg, and John MacCormick.
\newblock Learning {{Switching Linear Models}} of {{Human Motion}}.
\newblock In {\em Advances in {{Neural Information Processing Systems}}},
  volume~13. {MIT Press}, 2000.

\bibitem{pengAnimatableNeuralRadiance2021}
Sida Peng, Junting Dong, Qianqian Wang, Shangzhan Zhang, Qing Shuai, Xiaowei
  Zhou, and Hujun Bao.
\newblock Animatable {{Neural Radiance Fields}} for {{Modeling Dynamic Human
  Bodies}}.
\newblock In {\em 2021 {{IEEE}}/{{CVF International Conference}} on {{Computer
  Vision}} ({{ICCV}})}, pages 14294--14303, {Montreal, QC, Canada}, Oct. 2021.
  {IEEE}.

\bibitem{prabhu2021active}
Viraj Prabhu, Arjun Chandrasekaran, Kate Saenko, and Judy Hoffman.
\newblock Active domain adaptation via clustering uncertainty-weighted
  embeddings.
\newblock In {\em Proceedings of the IEEE/CVF International Conference on
  Computer Vision}, pages 8505--8514, 2021.

\bibitem{rempeHuMoR3DHuman2021}
Davis Rempe, Tolga Birdal, Aaron Hertzmann, Jimei Yang, Srinath Sridhar, and
  Leonidas~J. Guibas.
\newblock {{HuMoR}}: {{3D Human Motion Model}} for {{Robust Pose Estimation}}.
\newblock In {\em 2021 {{IEEE}}/{{CVF International Conference}} on {{Computer
  Vision}} ({{ICCV}})}, pages 11468--11479, {Montreal, QC, Canada}, Oct. 2021.
  {IEEE}.

\bibitem{roy2018deep}
Soumya Roy, Asim Unmesh, and Vinay~P Namboodiri.
\newblock Deep active learning for object detection.
\newblock In {\em BMVC}, page~91, 2018.

\bibitem{savvaSceneGrokInferringAction2014}
Manolis Savva, Angel~X. Chang, Pat Hanrahan, Matthew Fisher, and Matthias
  Nie{\ss}ner.
\newblock {{SceneGrok}}: Inferring action maps in {{3D}} environments.
\newblock {\em ACM Transactions on Graphics}, 33(6):212:1--212:10, Nov. 2014.

\bibitem{sigalHumanEvaSynchronizedVideo2010}
Leonid Sigal, Alexandru~O. Balan, and Michael~J. Black.
\newblock {{HumanEva}}: {{Synchronized Video}} and {{Motion Capture Dataset}}
  and {{Baseline Algorithm}} for {{Evaluation}} of {{Articulated Human
  Motion}}.
\newblock {\em International Journal of Computer Vision}, 87(1-2):4--27, Mar.
  2010.

\bibitem{sohnLearningStructuredOutput2015}
Kihyuk Sohn, Honglak Lee, and Xinchen Yan.
\newblock Learning {{Structured Output Representation}} using {{Deep
  Conditional Generative Models}}.
\newblock In {\em Advances in {{Neural Information Processing Systems}}},
  volume~28. {Curran Associates, Inc.}, 2015.

\bibitem{starkeNeuralStateMachine2019}
Sebastian Starke, He Zhang, Taku Komura, and Jun Saito.
\newblock Neural state machine for character-scene interactions.
\newblock {\em ACM Transactions on Graphics}, 38(6):209:1--209:14, Nov. 2019.

\bibitem{su2020active}
Jong-Chyi Su, Yi-Hsuan Tsai, Kihyuk Sohn, Buyu Liu, Subhransu Maji, and
  Manmohan Chandraker.
\newblock Active adversarial domain adaptation.
\newblock In {\em Proceedings of the IEEE/CVF Winter Conference on Applications
  of Computer Vision}, pages 739--748, 2020.

\bibitem{su2021anerf}
Shih-Yang Su, Frank Yu, Michael Zollh{\"o}fer, and Helge Rhodin.
\newblock A-{{NeRF}}: {{Articulated}} neural radiance fields for learning human
  shape, appearance, and pose.
\newblock In {\em Advances in Neural Information Processing Systems}, 2021.

\bibitem{tan2018}
Fuwen Tan, Crispin Bernier, Benjamin Cohen, Vicente Ordonez, and Connelly
  Barnes.
\newblock Where and who? {{Automatic}} semantic-aware person composition.
\newblock In {\em {{IEEE}} Winter Conf. on Applications of Computer Vision
  ({{WACV}})}, 2018.

\bibitem{urtasunTopologicallyconstrainedLatentVariable2008}
Raquel Urtasun, David~J. Fleet, Andreas Geiger, Jovan Popovi{\'c}, Trevor~J.
  Darrell, and Neil~D. Lawrence.
\newblock Topologically-constrained latent variable models.
\newblock In {\em Proceedings of the 25th International Conference on
  {{Machine}} Learning - {{ICML}} '08}, pages 1080--1087, {Helsinki, Finland},
  2008. {ACM Press}.

\bibitem{vondrick2011video}
Carl Vondrick and Deva Ramanan.
\newblock Video annotation and tracking with active learning.
\newblock {\em Advances in Neural Information Processing Systems}, 24, 2011.

\bibitem{wangDiverseNaturalSceneAware2022}
Jingbo Wang, Yu Rong, Jingyuan Liu, Sijie Yan, Dahua Lin, and Bo Dai.
\newblock Towards {{Diverse}} and {{Natural Scene-Aware 3D Human Motion
  Synthesis}}.
\newblock {\em CVPR}, page~10, 2022.

\bibitem{wangSynthesizingLongTerm3D2021}
Jiashun Wang, Huazhe Xu, Jingwei Xu, Sifei Liu, and Xiaolong Wang.
\newblock Synthesizing {{Long-Term 3D Human Motion}} and {{Interaction}} in
  {{3D Scenes}}.
\newblock In {\em 2021 {{IEEE}}/{{CVF Conference}} on {{Computer Vision}} and
  {{Pattern Recognition}} ({{CVPR}})}, pages 9396--9406, {Nashville, TN, USA},
  June 2021. {IEEE}.

\bibitem{wangGeometricPoseAffordance2021}
Zhe Wang, Liyan Chen, Shaurya Rathore, Daeyun Shin, and Charless Fowlkes.
\newblock Geometric {{Pose Affordance}}: {{3D Human Pose}} with {{Scene
  Constraints}}, Dec. 2021.

\bibitem{wangCriticRegularizedRegression2020}
Ziyu Wang, Alexander Novikov, Konrad {\.Z}o{\l}na, Jost~Tobias Springenberg,
  Scott Reed, Bobak Shahriari, Noah Siegel, Josh Merel, Caglar Gulcehre,
  Nicolas Heess, and Nando {de Freitas}.
\newblock Critic regularized regression.
\newblock In {\em Proceedings of the 34th {{International Conference}} on
  {{Neural Information Processing Systems}}}, {{NIPS}}'20, pages 7768--7778,
  {Red Hook, NY, USA}, Dec. 2020. {Curran Associates Inc.}

\bibitem{xiaLearningBasedSphereNonlinear2019}
Guiyu Xia, Huaijiang Sun, Qingshan Liu, and Renlong Hang.
\newblock Learning-{{Based Sphere Nonlinear Interpolation}} for {{Motion
  Synthesis}}.
\newblock {\em IEEE Transactions on Industrial Informatics}, 2019.

\bibitem{xuHierarchicalStylebasedNetworks2020}
Jingwei Xu, Huazhe Xu, Bingbing Ni, Xiaokang Yang, Xiaolong Wang, and Trevor
  Darrell.
\newblock Hierarchical {{Style-based Networks}} for {{Motion Synthesis}}.
\newblock {\em ECCV}, page~16, 2020.

\bibitem{yangViSERVideoSpecificSurface2021}
Gengshan Yang, Deqing Sun, Varun Jampani, Daniel Vlasic, Forrester Cole, Ce
  Liu, and Deva Ramanan.
\newblock {{ViSER}}: {{Video-Specific Surface Embeddings}} for {{Articulated 3D
  Shape Reconstruction}}.
\newblock In {\em Advances in {{Neural Information Processing Systems}}},
  volume~34, pages 19326--19338. {Curran Associates, Inc.}, 2021.

\bibitem{yangBANMoBuildingAnimatable2022}
Gengshan Yang, Minh Vo, Natalia Neverova, Deva Ramanan, Andrea Vedaldi, and
  Hanbyul Joo.
\newblock {{BANMo}}: {{Building Animatable 3D Neural Models From Many Casual
  Videos}}.
\newblock In {\em Proceedings of the {{IEEE}}/{{CVF Conference}} on {{Computer
  Vision}} and {{Pattern Recognition}}}, pages 2863--2873, 2022.

\bibitem{zhangPLACEProximityLearning2020}
Siwei Zhang, Yan Zhang, Qianli Ma, Michael~J. Black, and Siyu Tang.
\newblock {{PLACE}}: {{Proximity Learning}} of {{Articulation}} and {{Contact}}
  in {{3D Environments}}.
\newblock In {\em 2020 {{International Conference}} on {{3D Vision}}
  ({{3DV}})}, pages 642--651, {Fukuoka, Japan}, Nov. 2020. {IEEE}.

\bibitem{zhangGenerating3DPeople2020}
Yan Zhang, Mohamed Hassan, Heiko Neumann, Michael~J. Black, and Siyu Tang.
\newblock Generating {{3D People}} in {{Scenes Without People}}.
\newblock In {\em 2020 {{IEEE}}/{{CVF Conference}} on {{Computer Vision}} and
  {{Pattern Recognition}} ({{CVPR}})}, pages 6193--6203, {Seattle, WA, USA},
  June 2020. {IEEE}.

\bibitem{zhaoHumanNeRFEfficientlyGenerated2022}
Fuqiang Zhao, Wei Yang, Jiakai Zhang, Pei Lin, Yingliang Zhang, Jingyi Yu, and
  Lan Xu.
\newblock {{HumanNeRF}}: {{Efficiently Generated Human Radiance Field From
  Sparse Inputs}}.
\newblock {\em CVPR}, page~11, 2022.

\end{thebibliography}
}

\end{document}